\documentclass{article}
\usepackage{ijcai22}

\usepackage{times}
\usepackage{soul}
\usepackage{url}
\usepackage[hidelinks]{hyperref}
\usepackage[utf8]{inputenc}
\usepackage[small]{caption}
\usepackage{graphicx}
\usepackage{amsmath}
\usepackage{amsthm}
\usepackage{booktabs}
\usepackage{algorithm}
\usepackage{algorithmic}
\usepackage{amsfonts}
\usepackage{subfigure}
\usepackage{array}
\usepackage{tabu}
\usepackage{multirow}
\usepackage{diagbox}
\usepackage{xcolor}
\usepackage{lipsum}
\usepackage{bbding}
\usepackage{framed}
\usepackage{color}
\newcommand{\citet}[1] {\citeauthor{#1}~\shortcite{#1}}

\title{Parameter-Efficient Sparsity for Large Language Models Fine-Tuning}

\author{
Yuchao Li\and
Fuli Luo\and
Chuanqi Tan\and
Mengdi Wang\and \\
Songfang Huang\and
Shen Li\and
Junjie Bai
\affiliations
Alibaba Group
\emails
\{laiyin.lyc, lfl259702, chuanqi.tcq, didou.wmd, songfang.hsf, litan.ls, j.bai\}@alibaba-inc.com
}

\begin{document}

\maketitle

\begin{abstract}
  With the dramatically increased number of parameters in language models, sparsity methods have received ever-increasing research focus to compress and accelerate the models.
  While most research focuses on how to accurately retain appropriate weights while maintaining the performance of the compressed model, there are challenges in the computational overhead and memory footprint of sparse training when compressing large-scale language models.
  To address this problem, we propose a Parameter-efficient Sparse Training (PST) method to reduce the number of trainable parameters during sparse-aware training in downstream tasks.
  Specifically, we first combine the data-free and data-driven criteria to efficiently and accurately measure the importance of weights.
  Then we investigate the intrinsic redundancy of data-driven weight importance and derive two obvious characteristics \emph{i.e.}\ low-rankness and structuredness.
  Based on that, two groups of small matrices are introduced to compute the data-driven importance of weights, instead of using the original large importance score matrix, which therefore makes the sparse training resource-efficient and parameter-efficient.
  Experiments with diverse networks (\emph{i.e.} BERT, RoBERTa and GPT-2) on dozens of datasets demonstrate PST performs on par or better than previous sparsity methods, despite only training a small number of parameters.
  For instance, compared with previous sparsity methods, our PST only requires 1.5\% trainable parameters to achieve comparable performance on BERT.
  
\end{abstract}

\section{Introduction}
Many applications in natural language processing have been following a paradigm, which first pre-trains a large language model and then fine-tunes it towards multiple downstream tasks.
Despite its great success, such large-scale language models with millions to billions of parameters need a huge memory footprint and computational overhead in fine-tuning downstream datasets and also the inference stage, which prevents them from being directly applied to various tasks.

\begin{table}
  \small
\begin{center}
\begin{tabular}{p{1cm}<{\centering}p{2cm}<{\centering}p{1.5cm}<{\centering}p{2.5cm}<{\centering}}
\hline
\multirow{2}*{Method} & Extra Train Param. & \multirow{2}*{Need Data} & \multirow{2}*{Importance Criteria}\\
\hline

MaP & $0 \times$ & \XSolid & $|W|$ \\

MvP & $1 \times$ & \Checkmark & $-W*G$ \\

PST & $0.01 \sim 0.02 \times$ & \Checkmark & $|W|+AB+R+C$ \\

\hline
\end{tabular}
\end{center}
\caption{Comparison between different sparsity methods. MaP and MvP represent the representative data-free and data-driven methods, respectively. $W$ represents the weights, $G$ represents the corresponding gradient. $A$, $B$, $R$ and $C$ denote our proposed small matrices. We simplify the importance criteria for clear analysis.}
\label{tab:begin_compare}
\end{table}

To mitigate the computational and memory burden in the language model inference, 
one promising direction is pruning \cite{mccarley2019structured,zhang2020accelerating}, which removes unimportant weights/channels/layers independently to reduce the computation and memory overhead.
Among these, unstructured pruning, \emph{i.e.} sparsity, is widely studied since it can achieve a higher compression ratio with competitive performance.
Previous sparsity methods propose various criteria to compute the importance of each weight, which can be roughly classified to two categories, data-free \cite{han2015deep,tanaka2020pruning} and data-driven \cite{sanh2020movement,wang2020picking}.
The comparison is shown in Table~\ref{tab:begin_compare}.
Data-free criterion methods compute the importance of weight based on the weight itself without any data involved, such as magnitude pruning (MaP) \cite{han2015deep}.
Although data-free criteria have high computational and memory efficiency, they ignore that the role of each weight varies widely across different downstream tasks, which leads to degradation in model performance.
Typical data-driven criteria methods focus on designing precise important criteria to compute the importance scores based on the specific dataset, which is proved to succeed in reducing the computation inference cost of the language model without a performance drop.
However, these data-driven criteria introduce extra computation and trainable parameters to obtain the importance measurement, which dramatically increases the memory footprint and computational overhead during sparsity-aware training.
For example, movement pruning (MvP) \cite{sanh2020movement} computes the importance by multiplying the weights and their gradients and therefore needs extra memory to save importance scores matrix, which has the same size as the weights.
GraSP \cite{wang2020picking} introduces extra computational overhead to compute the hessian-gradient product.

\begin{figure}[t]
\includegraphics[width=1\columnwidth]{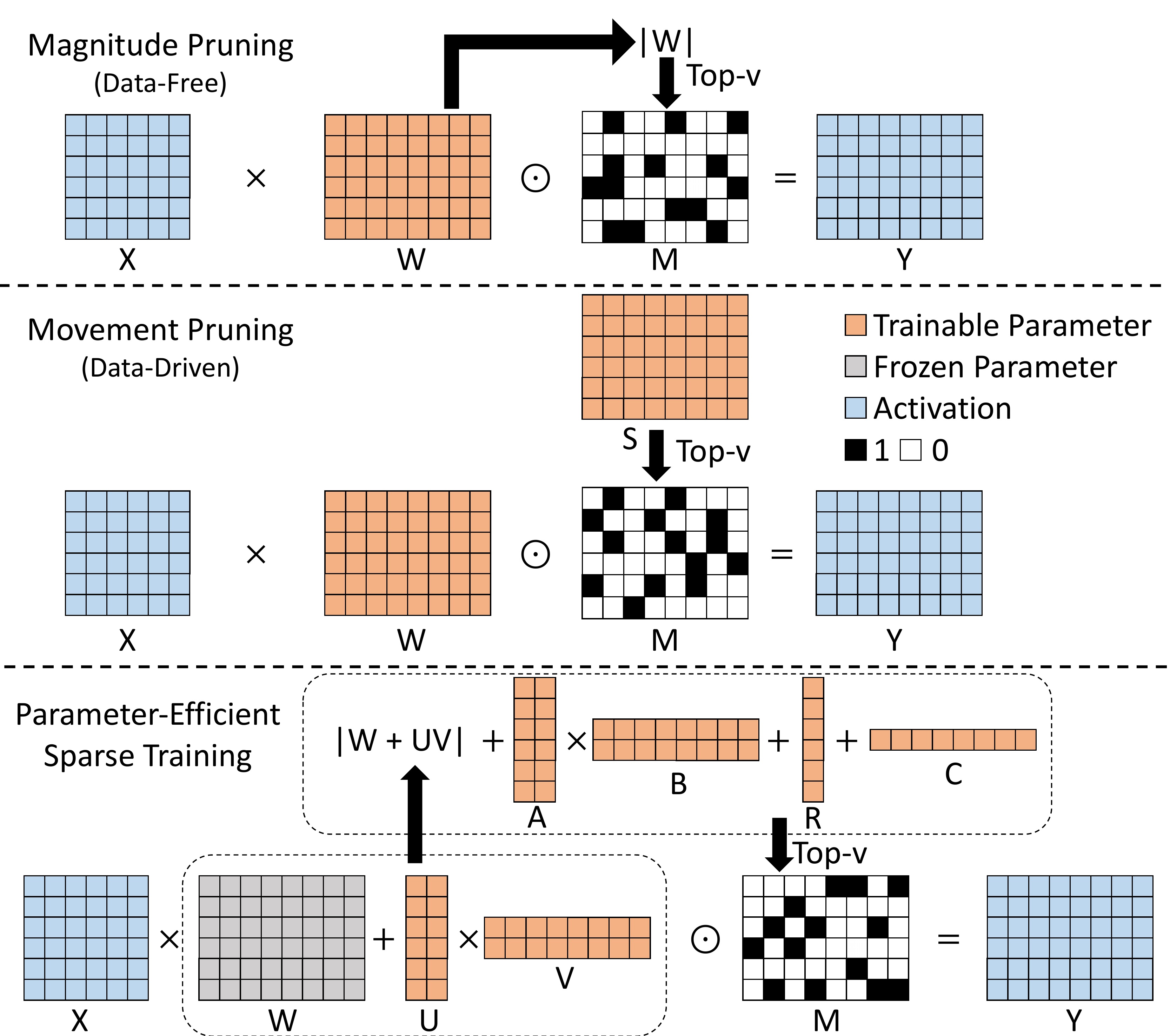}
 \caption{The framework of magnitude pruning, movement pruning, and our PST method. The magnitude pruning only optimizes the weight $W$, and the movement pruning simultaneously optimizes the weight $W$ and importance score $S$ to compute the sparse binary mask $M$. In our PST method, the update of weight is replaced by two small matrices ($U$ and $V$), and the data-driven importance score is decomposed into two sets of small matrices (\emph{i.e.,} $A$,$B$ and $R$,$C$) based on its low-rankness and structuredness.}
 \label{fig:framework}
  \end{figure}

In this paper, we propose a Parameter-efficient Sparse Training (PST) method to reduce the number of parameters involved in the weight importance computation, which can tackle the resource requirement issue in the sparse training while computing the accurate importance score.
Considering the efficiency of data-free criteria and the accurateness of data-driven criteria, the combination of them is adopted to leverage the advantage of both.
After that, to reduce the number of extra trainable parameters, \emph{i.e.} importance scores introduced by data-driven criteria, the training of the huge importance matrix is converted to the tuning of multiple small matrices, based on the two following basic observations,
\begin{itemize}
\item \textbf{Low-rankness:} we analyze the rank of weights and gradients based on previous works and observe that all of them have extremely low ranks, which means that the rank of the importance score matrix (combination of weight and gradient matrix) is also small. Therefore it can be represented by a set of rank-decomposition matrices (\emph{i.e.,} $A$ and $B$ in Table~\ref{tab:begin_compare} and Fig.~\ref{fig:framework}).
\item  \textbf{Structuredness:} we investigate the distribution of sparse weights and observe the phenomenon that there are some rows/columns less important than the others in general, which inspires us to introduce a set of small matrices to measure the importance of each row/column in weight. (\emph{i.e.,} $R$ and $C$ in Table~\ref{tab:begin_compare} and Fig.~\ref{fig:framework})
\end{itemize}

Two sets of small matrices are introduced to represent the low-rankness and structuredness in the data-driven importance scores, respectively.
The computation of importance scores in the specific downstream task is reformulated by these small matrices.
With the replacement, the resource requirement for data-driven criteria computation is dramatically reduced.
Moreover, we further reduce the number of trainable parameters by representing the update of weights with a low-rank decomposition, which optimizes a set of low-rank matrices instead of weight to capture the change of it.

Our contributions can be summarized as follows:
\begin{itemize}
\item We propose the Parameter-efficient Sparse Training (PST) method, which reduces the number of trainable parameters for the large language model sparse training and thus optimizes the fine-tuning and inference process in a parameter-efficient way.

\item
We exploit both the low-rankness and structuredness in the data-driven importance score and thus replace it with several small matrices. This leads to a novel research area, how to compress the redundancy of the importance score to efficiently obtain the importance of weights.

\item
Extensive experiments demonstrate the effectiveness of our method across various typical pre-trained large language models (\emph{e.g., } BERT, RoBERTa, and GPT-2) upon diverse datasets.
In particular, compared with previous works, PST obtains 98.5\% trainable parameter saving with a 0.12 average score improvement in GLUE.

\end{itemize}

\section{Related Works}

\textbf{Parameter-efficient fine-tuning.}
Parameter-efficient fine-tuning reduces the number of trainable parameters by optimizing various lightweight modules instead of original pre-trained weight.
For instance, \cite{houlsby2019parameter} introduced a trainable adapter with small number of parameters to achieve the parameter-efficient fine-tuning.
 \cite{lester2021power} proposed efficient prompt tuning which only optimized a small task-specific vector.
 \cite{he2021towards} presented a unified framework that employs multiple modules from previous works.
Besides, \cite{guo2020parameter} proposed only updating a small number of elements in the trainable vectors for parameter-efficient fine-tuning.
%;
\cite{hu2021lora} introduced two low-rank matrices to approximate parameter updates.
However, finetuned models produced by these methods have the same number of weight as the pre-trained model, which still leads to huge computation and memory overhead when inference. 
Different from them, we propose a parameter-efficient sparse training method to prune the unimportant weights in the language model during training, which reduces the resource requirement of network inference.

\noindent\textbf{Parameter-efficient inference.}
There are several popular language model compression techniques, \emph{e.g.,} pruning, quantization, and low-rank decomposition.
Among these, pruning is widely-used, which reduces the number of parameters in the network inference.
Structured pruning directly removes structured weights (\emph{e.g.}, attention heads \cite{mccarley2019structured}, channels \cite{wang2020structured} or layers \cite{zhang2020accelerating}) to compress and accelerate the large language models.
By contrast, unstructured pruning, \emph{i.e.} sparsity, removes the individual unimportant weights independently.
Previous works proposed various criteria to select insignificant weights for pruning, such as absolute weight \cite{gordon2020compressing}, taylor approximation \cite{molchanov2019importance}, hessian-gradient product \cite{wang2020picking} and data-free saliency scores \cite{tanaka2020pruning}.
However, these methods either propose a computation-efficient importance criterion but lead to worse network performance (\emph{i.e.,} magnitude pruning), or design an accurate importance criterion which may need huge computation overhead (\emph{i.e.,} movement pruning and GraSP).
Unlike these methods, our approach exploits intrinsic redundancy of the weight importance matrix and propose the parameter-efficient sparse training to obtain the better sparse network with lower resource requirement.

\section{Proposed Method}

\subsection{Preliminaries}

We first establish a general notation for analyzing the sparsity methods.
Generally, for a weight matrix $W \in \mathbb{R}^{n \times k}$, a network sparse strategy introduces an importance score $S \in \mathbb{R}^{n \times k}$ to determine which weights should be removed.
Based on $S$, a binary mask $M \in \{0, 1\}^{n \times k}$ can be generated for computation $Y = (W \odot M)X$, where $Y \in \mathbb{R}^{n \times m}$ and $X \in \mathbb{R}^{k \times m}$ are the output and input of the layer, respectively. $\odot$ denotes the Hadamard product.
A common strategy is to keep the top-$v$ of the weight $W$ based on the importance score $S$.
Thus, we define a function $f(S, v)$ which selects the $v$ largest values in $S$ to generate the binary mask $M$:
\begin{equation}
M_{i,j} = f(S, v)_{i,j} =
\left\{
\begin{aligned}
&1 , &S_{i,j} \text{in top-}v, \\
&0 , &\text{otherwise}.
\end{aligned}
\right.
\end{equation}
In this work, we focus on iterative sparse training, which removes the unimportant weights and updates the importance score step-by-step.
Previous methods prove that this strategy enables the network to recover from the information loss due to sparsity.
Thus, the optimized process of the language model fine-tuning is:
\begin{equation}
    \mathop{min}\limits_{W, S} \mathcal{L}(W \odot f(S, v); \mathcal{D}),~~ s.t. \frac{v}{n*k} \leq 1 - p
\end{equation}
\noindent where $\mathcal{D}$ is the observed dataset, $\mathcal{L}$ represents the loss function, and $p$ denotes the target compression ratio.
The update of $S$ depends on various sparse strategies. 
For example, movement pruning \cite{sanh2020movement} uses $S^{(t)} = -\sum\limits_{i=1}^{t} (\frac{\delta \mathcal{L}}{\delta W})^{(i)} \odot W^{(i)}$ to compute the importance score.

\subsection{Parameter-Efficient Sparse Training}

As presented in \cite{zhao2020masking} and \cite{zhang2021lottery}, the final binary mask generated by the trainable importance score is similar to that directly produced by the magnitude pruning, and the difference between them depends on the specific dataset.
It means that the importance of each weight depends on its absolute value and its role in the downstream tasks.
Thus, we propose a new importance score $S^{(t)} = |W^{(t)}| + \Delta S^{(t)}$, where $|W^{(t)}|$ and $\Delta S^{(t)}$ represent the data-free and data-driven importance of weight at the $t^{th}$-step, respectively.
Inspired by the works in \cite{sanh2020movement,zhang2021lottery}, we can directly optimize the importance score by SGD to obtain the data-driven importance score $\Delta S$, and thus the importance score at the $t^{th}$-step is re-written as:
\begin{equation}
    S^{(t)} = |W^{(t)}| - \alpha \sum\limits_{i=1}^{t} (\frac{\delta \mathcal{L}}{\delta W})^{(i)} \odot W^{(i)},
\end{equation}
\noindent where $\alpha$ is a hyper-parameter to trade-off the data-free and data-driven importance score.
For data-free importance score $|W^{(t)}|$, it does not need any extra parameters, which is resource-efficient.
Therefore, we only consider the compression of data-driven importance score $- \alpha \sum\limits_{i=1}^{t} (\frac{\delta \mathcal{L}}{\delta W})^{(i)} \odot W^{(i)}$ to achieve the parameter-efficient sparse training.

\textbf{Low-Rankness.}
As we known, $\text{rank}(W \odot \frac{\delta \mathcal{L}}{\delta W}) \leq \text{rank}(W) * \text{rank}(\frac{\delta \mathcal{L}}{\delta W})$, which means that the rank of data-driven importance score depends on the rank of $W$ and $\frac{\delta \mathcal{L}}{\delta W}$.
Previous work \cite{hu2021lora} proves that the gradient of weight $\frac{\delta \mathcal{L}}{\delta W}$ has a low intrinsic rank, which even can be one or two in the language models.
Thus the rank of the data-driven importance score matrix is close to the rank of the weight matrix.
Existing literature \cite{oymak2019generalization,li2021towards} shows that in the neural network, the trained large weight $W$ often naturally bears approximate low-rank weight structures.
According to that, we can derive the data-driven importance score also has a low intrinsic rank.
Thus, we introduce two small low-rank matrices $A \in \mathbb{R}^{n \times r_1}$ and $B \in \mathbb{R}^{r_1 \times k}$ to represent the low intrinsic rank part of data-driven importance score $\Delta S$, where $r_1$ is a hyper-parameter, controlling the number of trainable parameters for importance score.
To make the data-driven importance score of each weight the same at the beginning, $A$ and $B$ are initialized with Gaussian initialization and zero initialization respectively, and are directly optimized by SGD.

\begin{figure}[t]
 \subfigure[Attention Query Layer]{
 \label{fig:structuredness:attnq}
    \includegraphics[width=0.47\columnwidth]{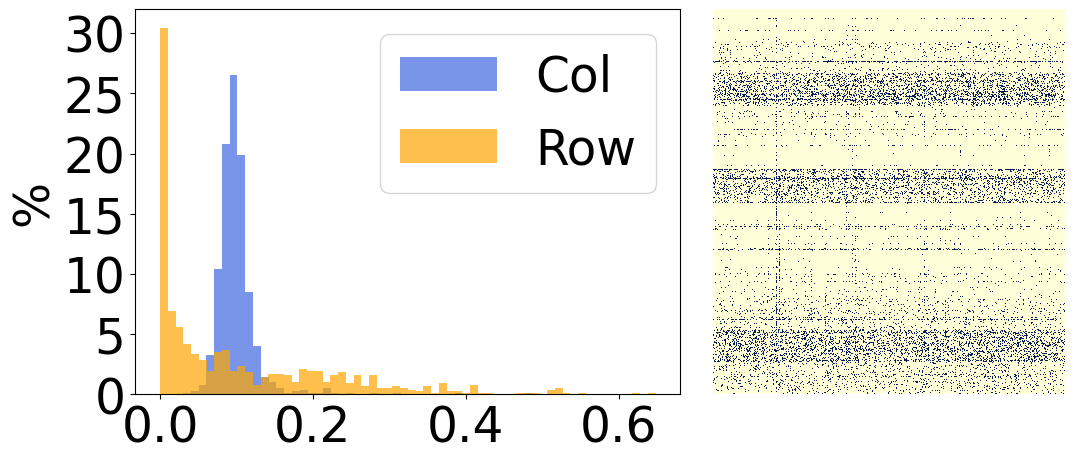}
    }
 \subfigure[Attention Output Layer]{
 \label{fig:structuredness:attno}
    \includegraphics[width=0.47\columnwidth]{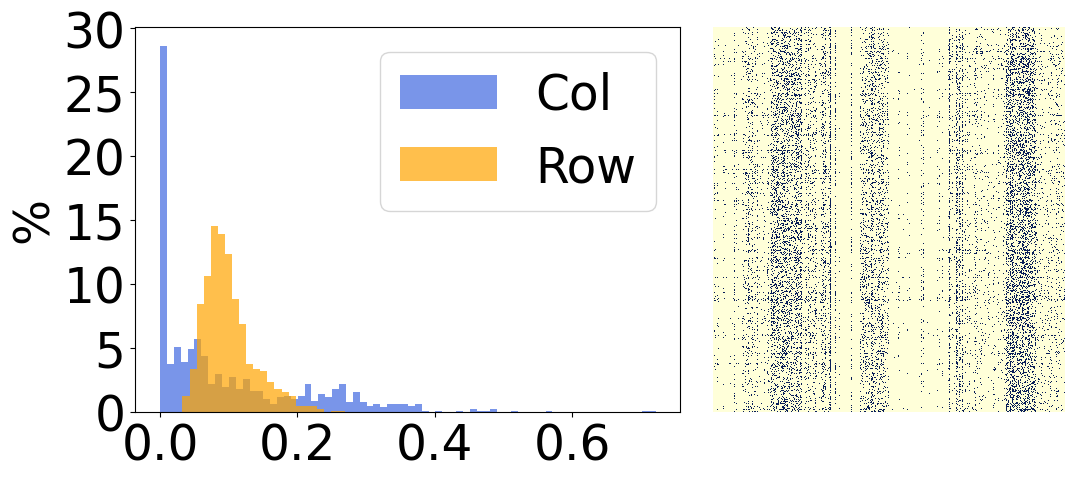}
  }	
  \subfigure[FFN Input Layer]{
    \includegraphics[width=0.47\columnwidth]{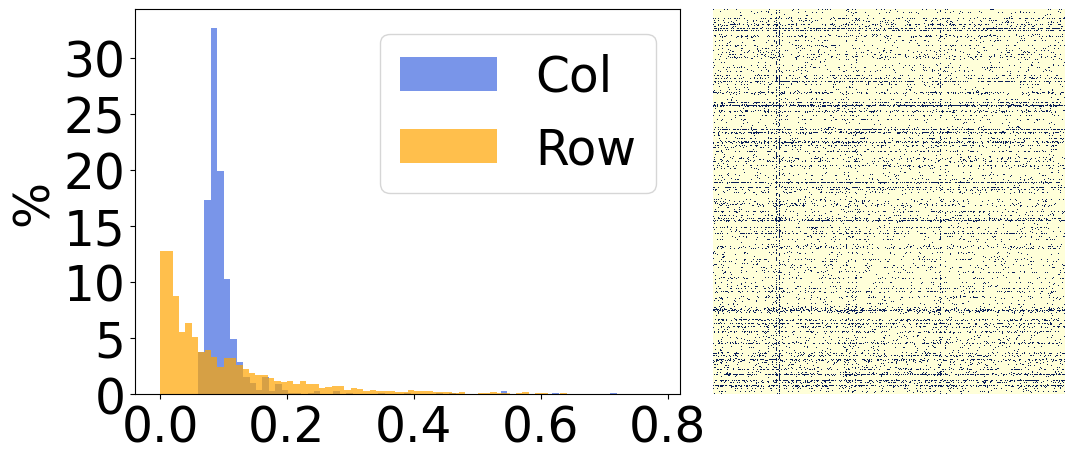}
  }	
  \subfigure[FFN Output Layer ]{
    \includegraphics[width=0.47\columnwidth]{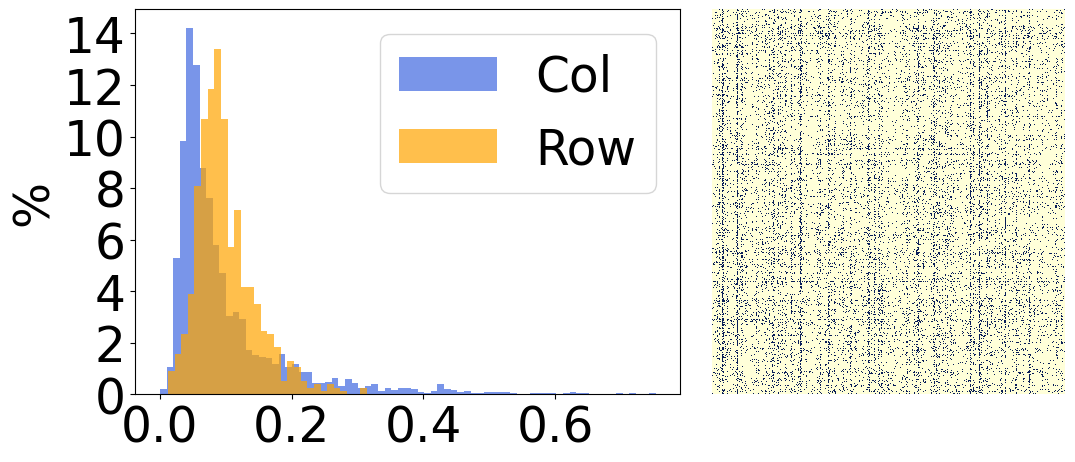}
  }	
 \caption{For each figure, the right sub-figure is the visualization of the binary mask $M$ in the first block of BERT on SST-2 when sparsity is 90\%. The left sub-figure is the corresponding sparsity distribution of column(blue) and row(orange). The x-axis represents the sparsity ratio and the y-axis represents the percentage of columns/rows whose sparsity ratio belongs to each interval.}
 \label{fig:structuredness}
  \end{figure}  

\textbf{Structuredness.}
Generally, sparsity methods remove the weights without any constraint, which means that the distribution of the sparse result (binary mask $M$) is uncontrollable.
However, as shown in Fig.~\ref{fig:structuredness}, the binary mask $M$ produced by importance score $S$ shows the obvious structural pattern.
For instance, the right sub-figure in Fig.~\ref{fig:structuredness:attnq} shows that there are many rows with extremely few weights reserved.
To quantify such a phenomenon, we compute the sparsity ratio of each column/row in binary $M$, then obtain their histograms by dividing the sparsity ratio into several intervals and computing the percentage of columns and rows whose sparsity ratios belong to corresponding intervals.
The left sub-figure in Fig.~\ref{fig:structuredness:attnq} demonstrates that there are about 30\% rows in which all weights are removed, while most columns have a similar sparsity ratio.
In contrast, Fig.~\ref{fig:structuredness:attno} shows that most columns have very high sparsity ratios.
Therefore, we conclude that the weights of the columns/rows differ significantly in importance.
Based on the observation, we propose two structural importance score matrices $R \in \mathbb{R}^{n \times 1}$ and $C \in \mathbb{R}^{1 \times k}$ to measure the importance of each column/row in the weight.
The update of them is:

\begin{equation}
\footnotesize
\begin{aligned}
    R^{(t)} = -\sum\limits_{i=0}^{t} \sum\limits_{j=0}^{k} [(\frac{\delta \mathcal{L}}{\delta W})^{(i)} \odot W^{(i)}]_{:, j}, \\
    C^{(t)} = -\sum\limits_{i=0}^{t} \sum\limits_{j=0}^{n} [(\frac{\delta \mathcal{L}}{\delta W})^{(i)} \odot W^{(i)}]_{j},
\end{aligned}
\end{equation}

\begin{table*}
  %\footnotesize
  \small
\begin{center}
\begin{tabular}{p{2cm}<{\centering}|p{1.5cm}<{\centering}|p{1.2cm}<{\centering}|>{\raggedleft\arraybackslash}p{1.2cm}|p{0.65cm}<{\centering} p{0.65cm}<{\centering}p{0.65cm}<{\centering}p{0.75cm}<{\centering}p{0.65cm}<{\centering}p{0.8cm}<{\centering}p{0.65cm}<{\centering}p{0.65cm}<{\centering}p{0.65cm}<{\centering}}
\hline
\multirow{2}*{Model} & \multirow{2}*{Method} & Sparsity Ratio & Trainable Param. & \multirow{2}*{MNLI} & \multirow{2}*{QQP} & \multirow{2}*{QNLI} & \multirow{2}*{SST-2} & \multirow{2}*{CoLA} & \multirow{2}*{STS-B} & \multirow{2}*{MRPC} & \multirow{2}*{RTE} & \multirow{2}*{Avg.} \\
\hline
\multirow{8}*{BERT$_{\text{base}}$} & Fine-tune & 0\% & 110.00M & 84.72 & 87.80 & 91.49 & 93.00 & 58.55 & 88.68 & 89.45 & 62.82 & 82.06 \\
\cline{2-13}
 & MaP & 50\% & 110.00M & \textbf{83.58} & \textbf{87.80} & \textbf{91.47} & 90.94 & \textbf{60.11} & \textbf{89.78} & 90.73 & 67.15 & \textbf{82.70} \\
% \cline{2-13}
& MvP & 50\% & 194.93M & 82.26 & 87.33 & 90.83 & 90.83 & 57.66 & 89.43 & \textbf{91.06} & 67.15 & 82.07 \\
% \cline{2-13}
& PST & 50\% & 2.91M & 80.97 & 85.77 & 89.77 & \textbf{91.28} & 57.60 & 84.63 & 90.72 & \textbf{67.87} & 81.08 \\

\cline{2-13}
 & MaP & 90\% & 110.00M & 79.75 & 82.83 & 85.06 & 87.04 & 40.74 & 81.72 & 82.78 & 54.87 & 74.35 \\
% \cline{2-13}
& MvP & 90\% & 194.93M & \textbf{80.06} & \textbf{85.37} & \textbf{86.53} & 87.04 & 40.46 & \textbf{84.35} & 84.28 & 58.84 & 75.87 \\
% \cline{2-13}
& $L_0$ Regu$^{*}$ & 90\% & 194.93M & 77.90 & 81.90 & - & - & - & - & - & - & - \\
& PST & 90\% & 2.91M & 76.73 & 83.93 & 86.03 & \textbf{88.65} & \textbf{42.49} & 81.70 & \textbf{85.57} & \textbf{62.82} & \textbf{75.99} \\
\hline

\multirow{4}*{RoBERTa$_{\text{base}}$} & Fine-tune$^{*}$ & 0\% & 125.00M & 87.60 & 91.90 & 92.80 & 94.80 & 63.60 & 91.20 & 90.20 & 78.70 & 86.40 \\
\cline{2-13}
 & MaP & 90\% & 125.00M & 80.85 & 84.90 & 85.70 & 88.99 & 19.13 & 83.58 & 83.82 & 55.23 & 72.78  \\
% \cline{2-13}
& MvP & 90\% & 209.93M & \textbf{81.40} & \textbf{86.42} & 87.13 & 89.68 & \textbf{38.12} & \textbf{85.85} & 85.71 & 56.32 & \textbf{76.33}  \\
% \cline{2-13}
& PST & 90\% & 2.91M & 76.70 & 83.83 & \textbf{87.26} & \textbf{90.02} & 38.08 & 84.94 & \textbf{87.34} & \textbf{60.29} & 76.06  \\
\hline

\multirow{4}*{RoBERTa$_{\text{large}}$} & Fine-tune$^{*}$ & 0\% & 355.00M & 90.20 & 92.20 & 94.70 & 96.40 & 68.00 & 92.40 & 90.90 & 86.60 & 88.90 \\
\cline{2-13}
 & MaP & 90\% & 355.00M & 79.37 & 83.29 & 85.83 & 89.68 & 14.94 & 80.21 & 82.77 & 58.12 & 71.78 \\
% \cline{2-13}
& MvP & 90\% & 682.36M & \textbf{82.91} & \textbf{85.94} & \textbf{88.27} & 90.83 & 32.50 & 84.20 & 85.20 & \textbf{59.93} & 76.22 \\
% \cline{2-13}
& PST & 90\% & 7.77M & 81.40 & 85.21 & 87.64 & \textbf{90.83} & \textbf{39.29} & \textbf{84.95} & \textbf{87.07} & 59.21 & \textbf{76.95} \\
\hline
\end{tabular}
\end{center}
\caption{Results of different network sparsity methods with BERT$_{\text{base}}$ and RoBERTa$_{\text{large}}$ on the GLUE benchmark. $*$ indicates numbers published in prior works. Bold number represents the best results under the same sparsity ratio.}
\label{tab:bert_acc}
\end{table*}

\begin{table*}
  \small
\begin{center}
\begin{tabular}{p{1.3cm}<{\centering}|p{1cm}<{\centering}|>{\raggedleft\arraybackslash}p{1.5cm}|p{1cm}<{\centering}p{1cm}<{\centering}p{1cm}<{\centering}|p{1cm}<{\centering}p{1cm}<{\centering}p{1cm}<{\centering}|p{1cm}<{\centering}p{1cm}<{\centering}p{1cm}<{\centering}}
\hline
\multirow{2}*{Method} & Sparsity & Trainable & \multicolumn{3}{c|}{E2E} & \multicolumn{3}{c|}{DART} & \multicolumn{3}{c}{WebNLG} \\
 & Ratio & Param. & BLEU & MET & NIST & BLEU & MET & TER & BLEU & MET & TER \\
\hline
 Fine-tune & 0\% & 354.92M & 68.36 & 46.41 & 8.66 & 46.00 & 0.39 & 0.46 & 47.60 & 0.39 & 0.50 \\

 MaP & 90\% & 354.92M & 68.42 & 46.08 & 8.64 & 44.72 & 0.37 & 0.50 & 37.38 & 0.30 & 0.64 \\

 MvP & 90\% & 656.91M & 69.24 & 46.36 & 8.73 & 45.11 & 0.37 & 0.50 & 38.32 & 0.32 & 0.63 \\

 PST & 90\% & 7.77M & \textbf{70.04} & \textbf{46.51} & \textbf{8.81} & \textbf{45.27} & \textbf{0.37} & \textbf{0.49} & \textbf{44.57} & \textbf{0.34} & \textbf{0.53} \\
\hline

\end{tabular}
\end{center}
\caption{GPT-2 medium performance on E2E, DART and WebNLG with different methods. For all metrics except TER, higher is better.}
\label{tab:gpt2}
\end{table*}

In summary, the data-driven importance score becomes:
\begin{equation}
    \Delta S^{(t)} = \alpha_1 A^{(t)} B^{(t)} + \alpha_2 (R^{(t)} + C^{(t)}), 
\end{equation}
\noindent where the $\alpha_1$ and $\alpha_2$ are the hyper-parameters to trade-off the low-rankness and structural importance score, respectively.

To further reduce the resource-requirement of the sparse training, we follow \cite{hu2021lora} to constrain the update of weight by representing it with a low-rank decomposition $W^{(t)} = W^{(0)} + \beta U^{(t)} V^{(t)}$, where $U \in \mathbb{R}^{n \times r_2}$, $V \in \mathbb{R}^{r_2 \times k}$ and $r_2$ controls the trainable parameters of weight.
Therefore, the importance score in our method is:
\begin{equation}
\small
    S^{(t)} = |W^{(0)} + \beta U^{(t)} V^{(t)}| + \alpha_1 A^{(t)} B^{(t)} + \alpha_2 (R^{(t)} + C^{(t)}).
\end{equation}
Based on that, the computation of each layer becomes:
\begin{equation}
\small
\begin{aligned}
    Y = & [(W^{(0)} + \beta U^{(t)} V^{(t)}) \odot f(|W^{(0)} + \beta U^{(t)} V^{(t)}| \\ 
        & + \alpha_1 A^{(t)} B^{(t)} + \alpha_2 (R^{(t)} + C^{(t)}), v)] X.
\end{aligned}
\end{equation}

It should be noted that, after fine-tuning, all weights are finalized and the inference procedure will be $Y=W^*X$, where $W^*$ is sparse, $W^*=[(W^{(0)} + \beta U^{(t)} V^{(t)}) \odot f(|W^{(0)} + \beta U^{(t)} V^{(t)}| + \alpha_1 A^{(t)} B^{(t)} + \alpha_2 (R^{(t)} + C^{(t)}), v)]$. Therefore, the inference procedure is parameter- and resource-efficient.

The optimized process of our sparse training is:
\begin{equation}
\footnotesize
\begin{aligned}
    \mathop{min}\limits_{U, V, A, B, R, C} &\mathcal{L}((W^{(0)} + \beta U V) \odot f(|W^{(0)} + \beta U V|  \\ 
        & + \underbrace{\alpha_1 A B}_{Low-Rankness} + \underbrace{\alpha_2 (R + C)}_{Structuredness}, v); \mathcal{D}), \\
        & s.t.\ \frac{v}{n*k} \leq 1 - p
\end{aligned}
\end{equation}
In addition, the number of trainable parameters in our method is $(n + k) * (r_1 + r_2 + 1)$, which is extremely smaller than the original number $2 * n * k$ when $r_1$ and $r_2$ is small.

\section{Experiments}

\subsection{Evaluation Setup}
\noindent \textbf{Datasets and Backbone Models.}
We conduct experiments with BERT \cite{devlin2019bert}, RoBERTa \cite{liu2019roberta}, and GPT-2 \cite{radford2019language} in various downstream tasks.
For BERT and RoBERTa, we use GLUE benchmarks \cite{wang2018glue} for evaluation.
For GPT-2, we evaluate it on the E2E, DART, and WebNLG.

\noindent \textbf{Implementation Details.}
For BERT$_{\text{base}}$, we set batch size = $32$ and perform a hyperparameter search over learning rate $\in$ \{3e-5, 5e-5, 1e-4, 5e-4\} and epoch $\in \{20, 40\}$ on QNLI, SST-2, CoLA, STS-B, MRPC, RTE and epoch $\in \{10, 20\}$ on MNLI, QQP.
Moreover, we use a batch size of 16 for RoBERTa, as well as a hyperparameter search over learning rate $\in$ \{1e-5, 2e-5, 3e-5, 5e-5\}. Epoch search space is the same as BERT$_{\text{base}}$.
For GPT-2, we train the model for 5 epochs using a batch size of 8 and an initial learning rate of 1e-4.
At training time, we use the AdamW optimizer and a linear learning rate scheduler.
All models are initialized with the pre-trained weights.
We follow the \cite{zhu2018prune} to use a cubic sparsity scheduling.
We also add a few steps of warm-up at the beginning of training (10\% training steps) and cool-down at the end of training (30\% training steps), which empirically improve the performance especially in high sparsity regimes.
For PST, we set $\beta = \alpha_1 = \alpha_2 = 1$ and $r_1 = r_2 = 8$.\footnote{Our code is available at \url{https://github.com/alibaba/AliceMind/tree/main/S4/PST} and \url{https://github.com/yuchaoli/PST}.}
%

% \noindent \textbf{Baseline.}
% %
% We compare our methods against several representative and state-of-the-art sparsity methods: magnitude pruning (MaP) that only uses data-free importance scores, movement pruning (MvP) that uses data-driven importance scores, and $L_0$ regularization baseline reported in \cite{sanh2020movement}.
% %
% We observe that movement pruning will obtain terrible performance when initializing importance score with a constant value, so we initialize the importance score by the absolute value of the pre-trained weights for movement pruning.

\begin{figure}[t]
 \subfigure[MRPC]{
 \label{fig:sparsity:mrpc}
    \includegraphics[width=0.47\columnwidth]{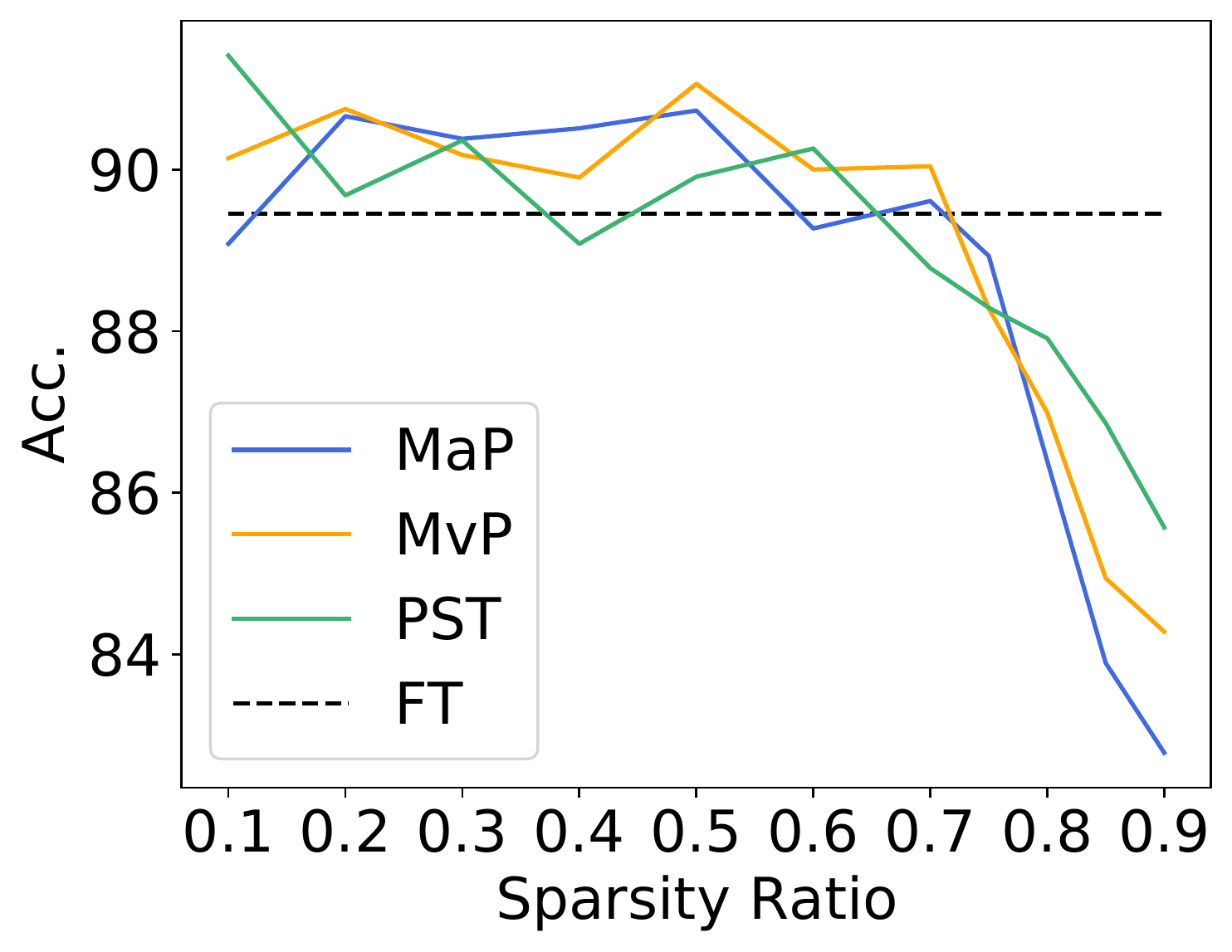}
    }
 \subfigure[SST-2]{
 \label{fig:sparsity:sst2}
    \includegraphics[width=0.47\columnwidth]{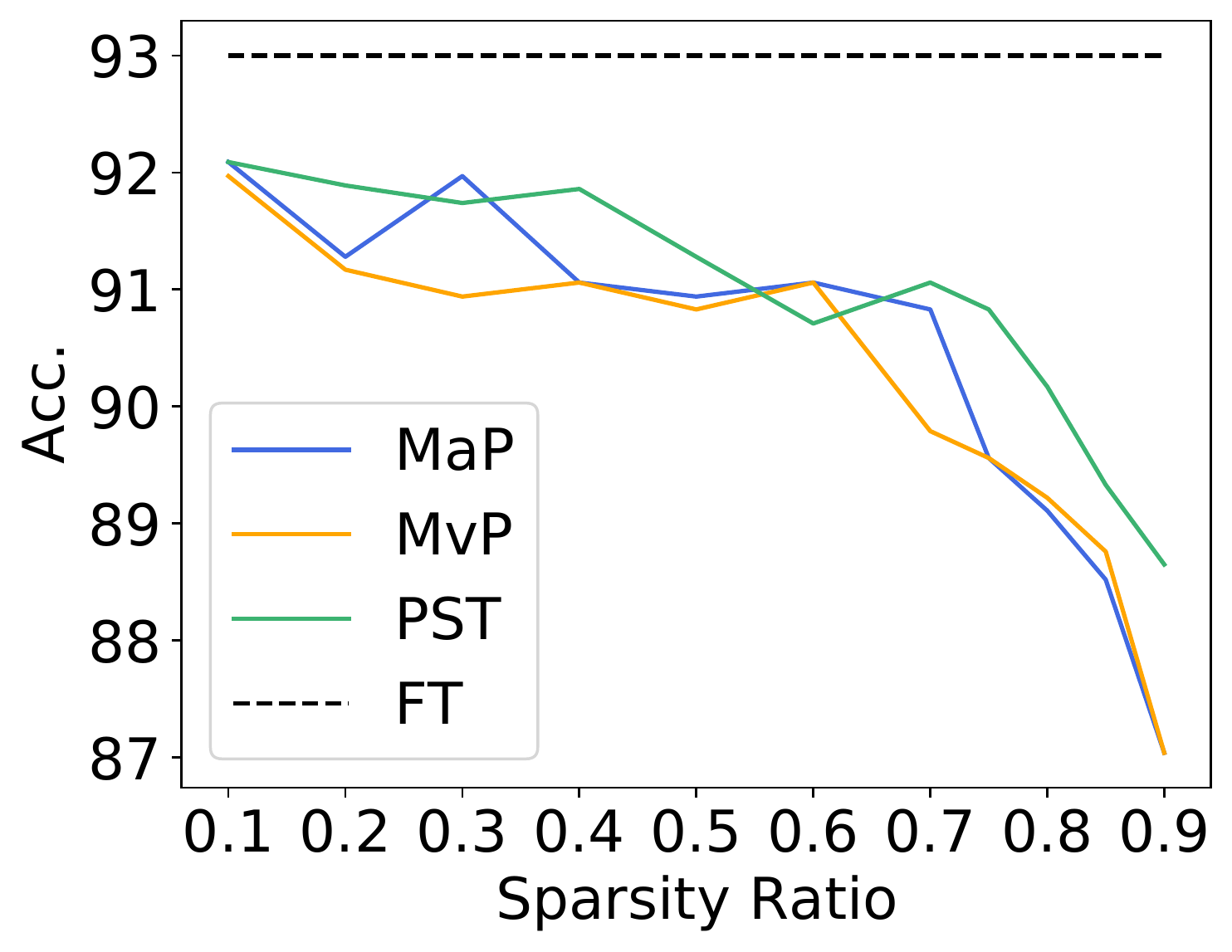}
  }	
 \caption{Comparison between different sparsity methods with different sparsity ratios on BERT$_{base}$.}
 \label{fig:sparsity}
\end{figure}

\subsection{Results}

\noindent \textbf{BERT and RoBERTa.}
Table~\ref{tab:bert_acc} shows that our method achieves the largest reduction of trainable parameters with on-par or better performance than previous methods.
We initialize the importance score by the absolute value of the pre-trained weights for movement pruning to avoid obtain terrible performance.
For instance, we achieve $0.73$ average score improvement with 98.9\% trainable parameter saving on RoBERTa$_{large}$ when the sparsity ratio is 90\%.
Moreover, we observe that MaP outperforms other methods with little or no loss with respect to the fine-tuned dense model at the low sparsity ratio (50\%).
However, when increasing the sparsity ratio to 90\%, it obtains an obvious performance drop whether in BERT or RoBERTa.
In contrast, our method PST performs poorly with the low sparsity ratio but obtains better performance than other methods at a higher sparsity ratio, which is also shown in Fig.~\ref{fig:sparsity}.
Meanwhile, although RoBERTa achieves better performance than BERT after fine-tuning, the model after sparse training performs worse than BERT.
For this case, we find that RoBERTa has a smaller default learning rate than BERT on downstream tasks, which indicates that RoBERTa relies more on pre-trained weights than BERT. The sparsity methods make some weights become zeros. These weight changes in RoBERTa may have a greater impact on downstream tasks. We have to note that it is not a common phenomenon, the larger models are usually more stable than smaller models in the field of model compression \cite{li2020train}.

\noindent \textbf{GPT-2.}
We further verify that our method can also prevail on the NLG model.
As shown in Table~\ref{tab:gpt2}, our PST achieves the best performance while training an extremely smaller number of parameters in three downstream tasks.
In particular, compared with MvP, we obtain 6.25 BLEU improvement while saving 98.8\% trainable parameters on WebNLG.

\subsection{Ablation Study}

\begin{table}[t]
\small
\makebox[0pt][t]{\parbox{0.48\textwidth}{%
    \begin{minipage}[t]{0.24\textwidth}
    \begin{center}
        \begin{tabular}{p{0.60cm}<{\centering}|p{0.5cm}<{\centering}p{0.5cm}<{\centering}p{0.5cm}<{\centering}}
            \hline
            \diagbox[width=1cm]{$r_1$}{$r_2$} & 4 & 8 & 16 \\
            \hline
            4 & 84.07 & 84.88 & 85.52 \\
            
            8 & 85.86 & 85.57 & 85.76 \\
            
            16 & 86.45 & \textbf{86.75} & 86.21 \\
            \hline
        \end{tabular}
    \end{center}
    \begin{center}
        (a) MRPC
    \end{center}
    \end{minipage}
    \hfill
    \begin{minipage}[t]{0.24\textwidth}
    \begin{center}
        \begin{tabular}{p{0.60cm}<{\centering}|p{0.5cm}<{\centering}p{0.5cm}<{\centering}p{0.5cm}<{\centering}}
            \hline
            \diagbox[width=1cm]{$r_1$}{$r_2$} & 4 & 8 & 16 \\
            \hline
            4 & 88.42 & 88.53 & 88.76 \\

            8 & 88.65 & 88.65 & 88.53 \\
            
            16 & 88.76 & \textbf{88.99} & 87.96 \\
            \hline
        \end{tabular}
    \end{center}
    \begin{center}
        (b) SST-2
    \end{center}
    \end{minipage}
}}
\caption{Comparison on BERT$_{\text{base}}$ with different rank $r_1$ and $r_2$.}
\label{tab:rank}
\end{table}

\begin{table*}
  \small
\begin{center}
\begin{tabular}{p{7cm}|p{1cm}<{\centering}p{1cm}<{\centering}p{1cm}<{\centering}p{1cm}<{\centering}p{1cm}<{\centering}p{1cm}<{\centering}p{1cm}<{\centering}}
\hline
$S$ (Importance Score)  & QNLI & SST-2 & CoLA & STS-B & MRPC & RTE & Avg. \\
\hline

$|W^{(0)} + \beta U V| + \alpha_1  A B + \alpha_2( R + C)$ & \textbf{86.03} & \textbf{88.65} & \textbf{42.49} & \textbf{81.7} & \textbf{85.57} & \textbf{62.82} & \textbf{74.54} \\

$|W^{(0)} + \beta UV| + \alpha_1  A B$ & 85.61 & 88.42 & 32.60 & 78.80 & 83.44 & 61.01 & 71.65 \\

$|W^{(0)} + \beta UV| + \alpha_2( R + C)$ & 85.58 & 88.19 & 37.71 & 81.67 & 85.34 & 62.82 & 73.55 \\

$|W^{(0)} + \beta UV|$ & 85.83 & 88.19 & 37.66 & 80.08 & 84.96 & 61.37 & 73.02 \\

\hline

$\alpha_1  A B + \alpha_2( R + C)$ & 85.48 & 87.50 & 32.90 & 80.52 & 84.95 & 62.82 & 72.36 \\

$\alpha_1  A B$ & 83.56 & 84.63 & 22.02 & 69.84 & 81.66 & 54.15 & 65.98 \\

$\alpha_2( R + C)$ & 85.10 & 87.27 & 34.93 & 81.50 & 85.12 & 61.73 & 72.61 \\

\hline
\end{tabular}
\end{center}
\caption{Comparison on BERT$_{\text{base}}$ of different importance scores with same number of trainable parameters ($p$ = 90\%).}
\label{tab:importance_score}
\end{table*}

\noindent \textbf{Importance score.}
The design of importance score plays a crucial role in our proposed PST.
We combine the data-free and data-driven importance score, and decompose data-driven importance score into two sets of small matrices based on its low-rankness and structuredness.
Precisely, we compare seven different importance scores on BERT$_{\text{base}}$ in Table~\ref{tab:importance_score}.
We adjust the $r_1$ and $r_2$ to make all of the methods have the same number of trainable parameters.
The results show that the proposed importance score achieves the best performance in various downstream tasks.
Furthermore, structuredness is more important than low-rankness for importance score compared with line 2 and 3.

\noindent \textbf{Rank $r_1$ and $r_2$.}
Table~\ref{tab:rank} shows the effect of the rank $r_1$ and $r_2$.
We observe that although the model performance increases as the rank increases, higher is not necessarily better.
When the one rank is lower (\emph{i.e.,} $r_1=4$ or $r_2=4$), another rank increases will improve the model accuracy.
But when the one rank is large enough (\emph{i.e.,} $r_1=16$ or $r_2=16$), the increase of another one does not necessarily improve the model performance.
This suggests that the rank $r_1$ and $r_2$ can also be searched to explore the most suitable configuration for different downstream tasks.

\begin{figure}[t]
 \subfigure[Distribution of sparse weights]{
 \label{fig:dis:weight}
    \includegraphics[width=0.47\columnwidth]{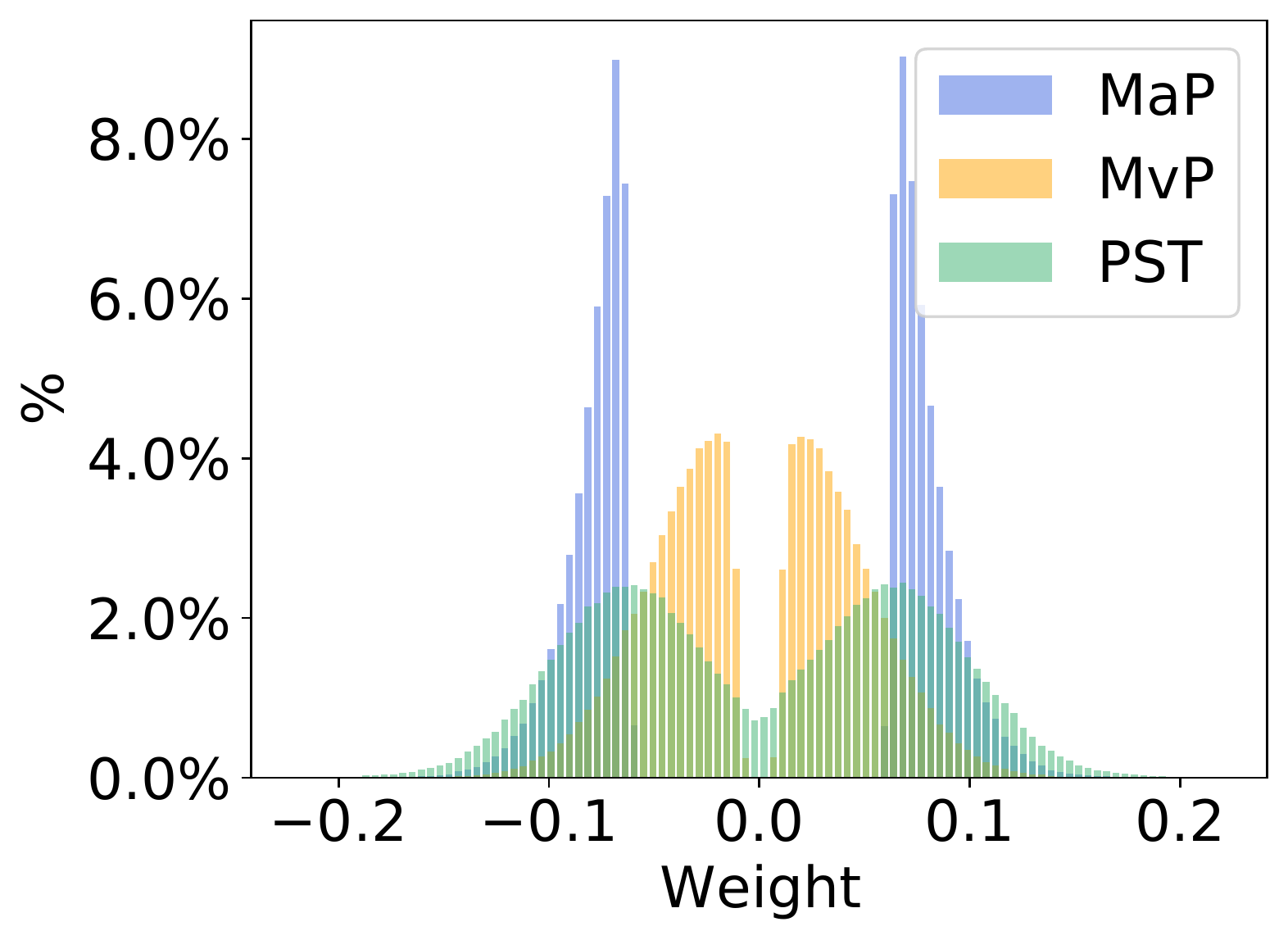}
    }
 \subfigure[Scores and weights in MaP]{
 \label{fig:dis:map}
    \includegraphics[width=0.47\columnwidth]{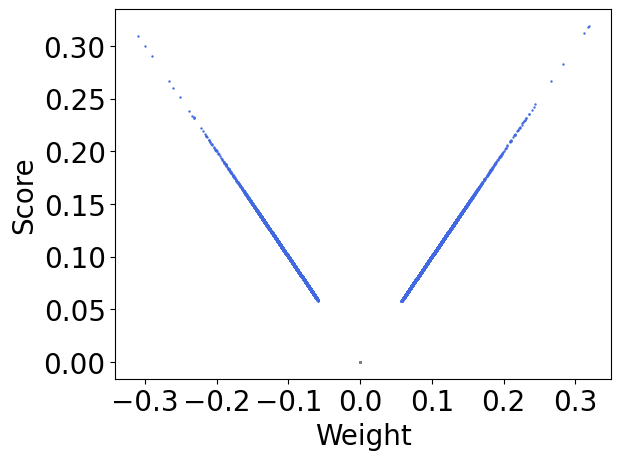}
  }	
  \subfigure[Scores and weights in MvP]{
 \label{fig:dis:mvp}
    \includegraphics[width=0.47\columnwidth]{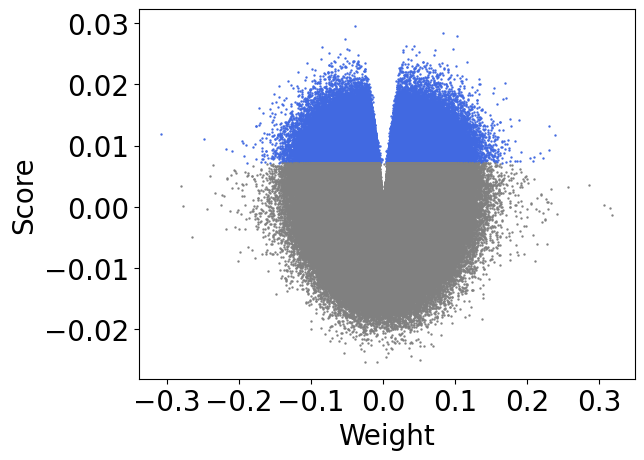}
  }	
  \subfigure[Scores and weights in PST]{
 \label{fig:dis:pst}
    \includegraphics[width=0.47\columnwidth]{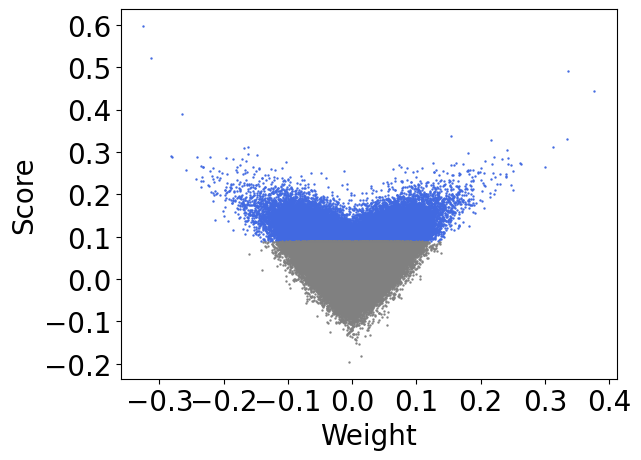}
  }	
%  \caption{The distribution of weights in sparse models based on MaP, MvP and PST.}
 \caption{Distribution of sparse weights of MaP, MvP and PST, respectively ($p$ = 90\%).}
 \label{fig:dis}
\end{figure}  

\begin{figure}[t]
 \subfigure[Attention Query Layer]{
 \label{fig:similar:attnq}
    \includegraphics[width=0.47\columnwidth]{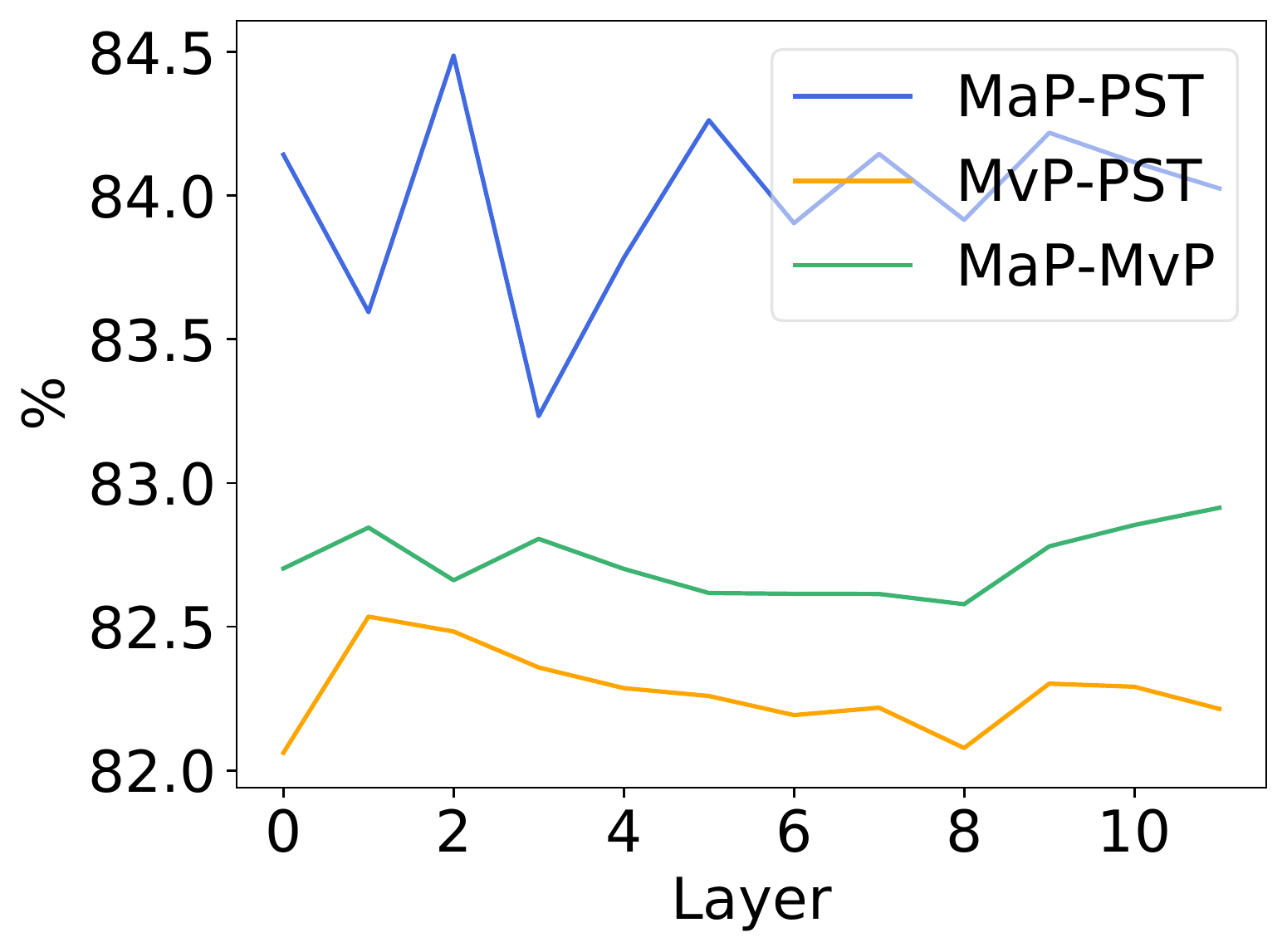}
    }
 \subfigure[Attention Output Layer]{
 \label{fig:similar:attno}
    \includegraphics[width=0.47\columnwidth]{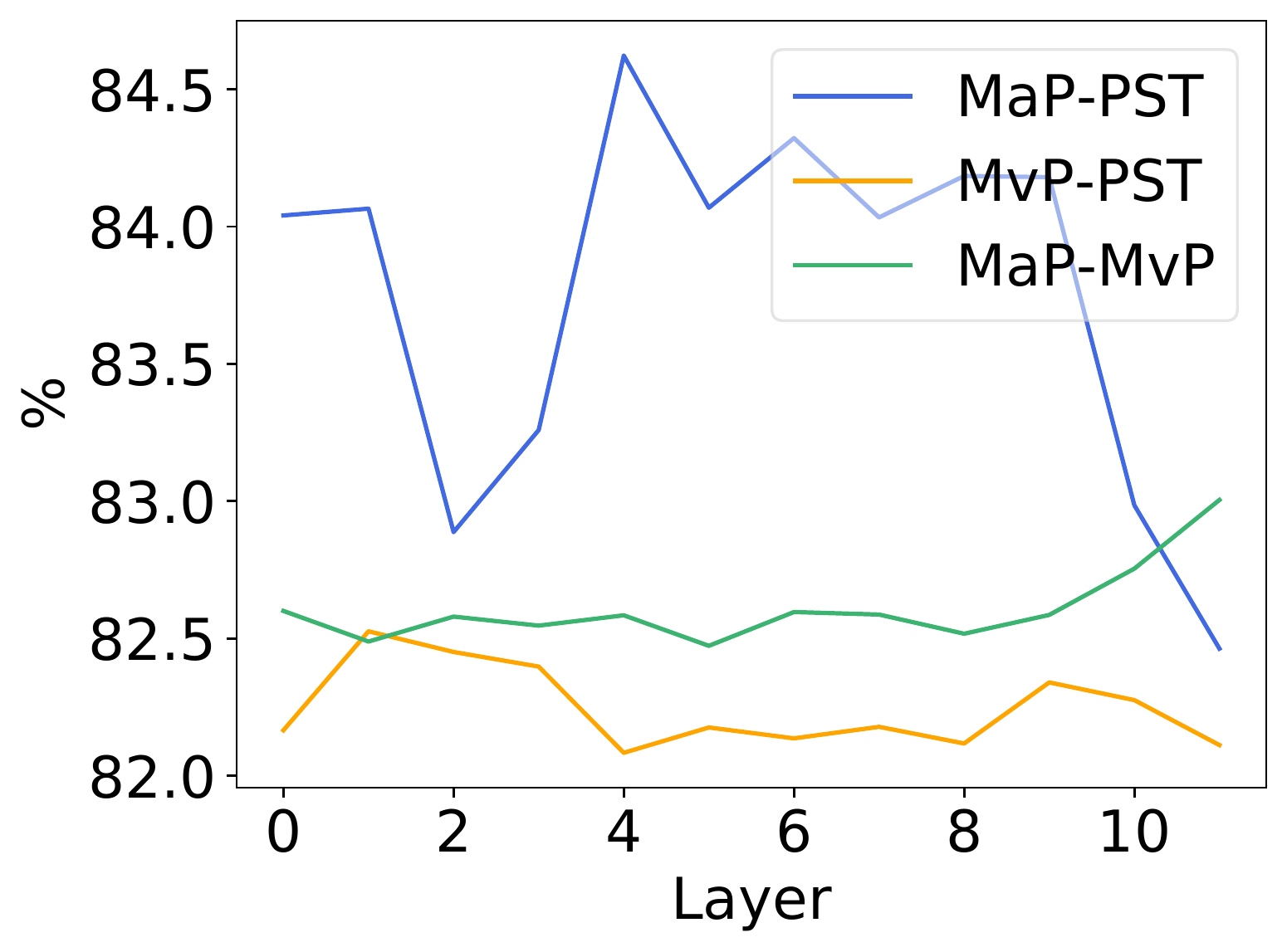}
  }	
  \subfigure[FFN Input Layer]{
 \label{fig:similar:ffni}
    \includegraphics[width=0.47\columnwidth]{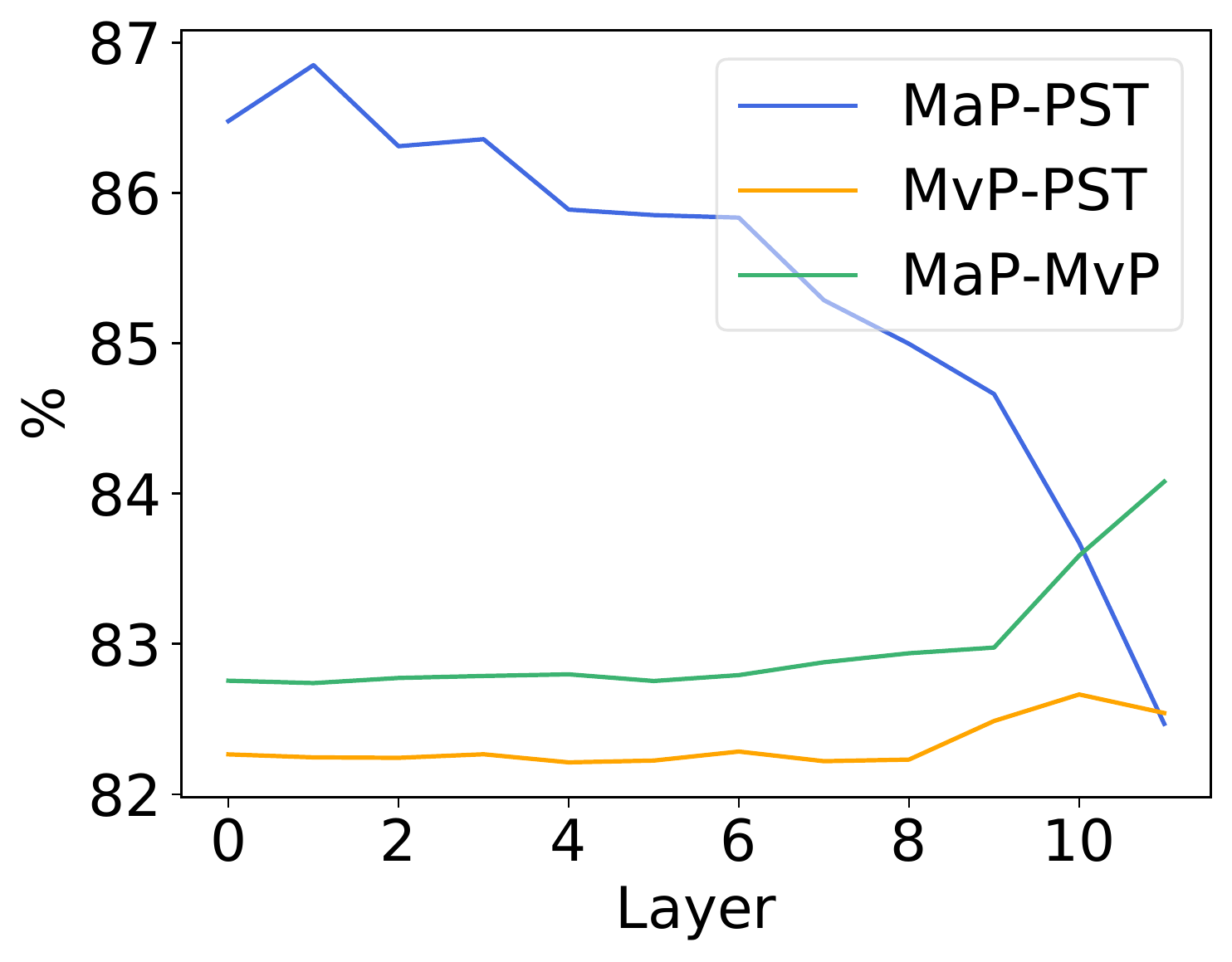}
  }	
  \subfigure[FFN Output Layer]{
 \label{fig:similar:ffno}
    \includegraphics[width=0.47\columnwidth]{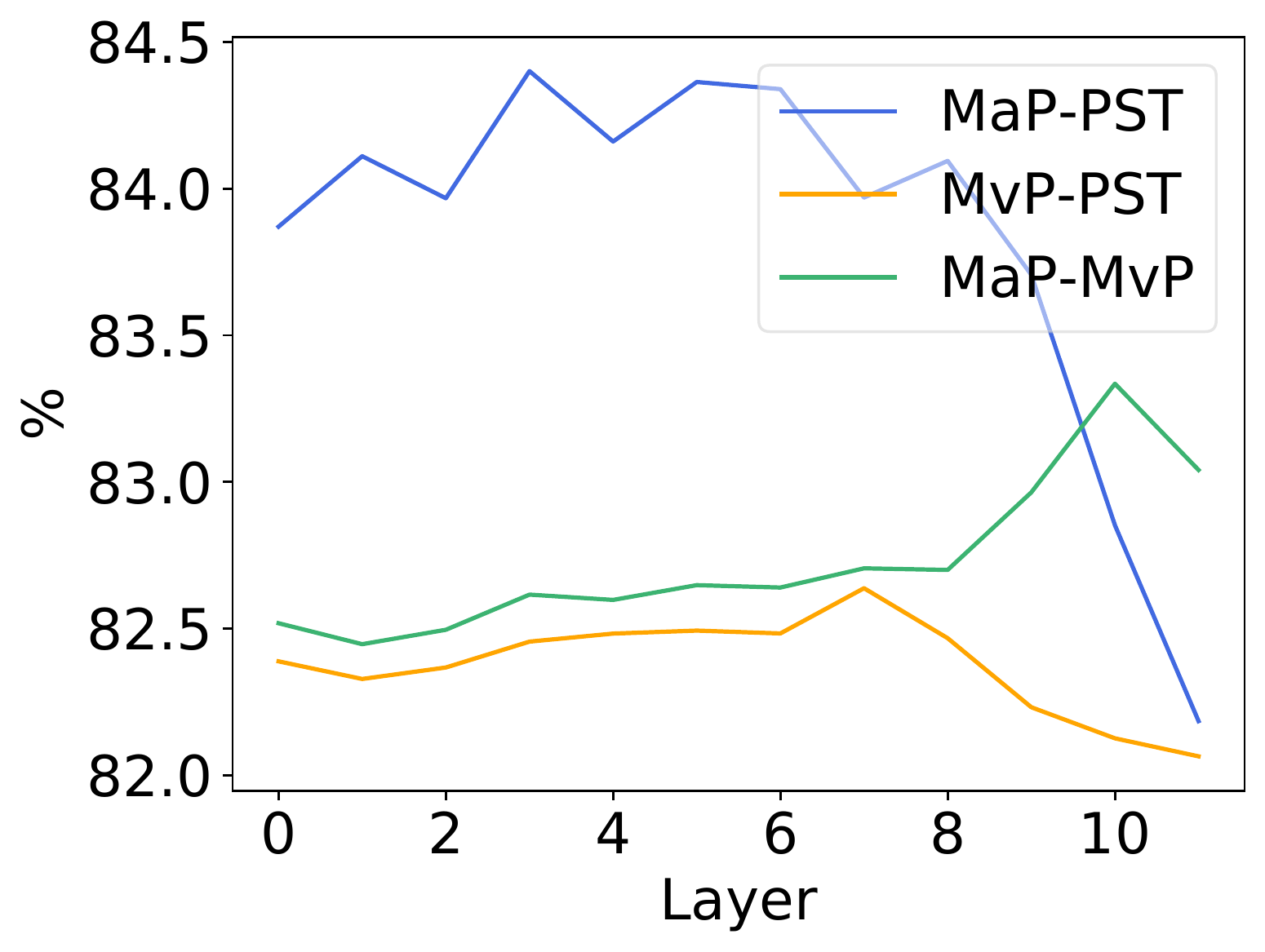}
  }	
 \caption{Similarity of the binary mask $M$ between MaP, MvP and PST, respectively ($p$ = 90\%).}
 \label{fig:similar}
\end{figure}  

\subsection{Analysis}

\noindent \textbf{Distribution of sparse weights.}
Fig.~\ref{fig:dis:weight} shows an overview of the distribution of the remaining weights of MaP, MvP and PST respectively at the same layer with a sparsity ratio of 90\%.
Compared with MaP that tends to remove weights close to zero and MvP that removes weights with the larger values, PST has a smoother distribution, which holds weights both with larger and smaller values.
Fig.~\ref{fig:dis}(b)(c)(d) display the weight against the importance score of MaP, MvP, and PST, respectively.
The pruned and remaining weights are grey and blue dot respectively.
We observe that the PST reflects the characteristics of both the data-free (MaP) and data-driven (MvP) methods.
MaP computes the importance score of weights based on their absolute values and thus shows a v-shaped curve.
MvP removes any weights regardless of their absolute values (except zero).
However, PST not only considers the absolute value of weight but also remains the weight with a low absolute value, and therefore shows a combination of their two distributions.

\noindent \textbf{Similarity of binary mask.}
We use the Hamming distance to compute the similarity of binary mask $M$ among different methods.
Fig.~\ref{fig:similar} shows that the sparse binary mask $M$ of PST is closer to MaP than MvP, which means that the data-free importance score accounts for a greater proportion in PST. 
%\
Moreover, as shown in Fig.~\ref{fig:similar:ffni} and Fig.~\ref{fig:similar:ffno}, the similarity between MaP and PST decreases when the depth of layers in the FFN module increases.
It demonstrates that the PST gradually reduces the impact of data-free importance score with the deepening of the layer.
However, with the increase of the depth of layers, the similarity between MvP and PST increases in the input layer of the FFN module and decreases in the output layer of the FFN module.
It indicates that the importance score of PST explores the new information that is different from MaP and MvP in the output layer.

\section{Conclusion}
In this paper, we propose a parameter-efficient sparse training (PST) method to reduce the number of trainable parameters and the resource requirements during sparse-aware fine-tuning of large language models.
We first combine the data-free and data-driven criteria to compute the importance of weights. Then we discover two characteristics (\emph{i.e.,} low-rankness and structuredness) of data-driven importance score, and therefore introduce two sets of parameter-efficient matrices to replace the original large importance score matrix.
Extensive experiments on various language models demonstrate the effectiveness of PST in reducing the computational complexity and resource requirements in sparse fine-tuning.

\bibliographystyle{named}
\bibliography{ijcai22}

\end{document}